\RequirePackage{amsmath}
\documentclass[runningheads]{llncs}
\usepackage[T1]{fontenc}

\usepackage{graphicx}
\usepackage{textcomp}
\usepackage{gensymb}
\usepackage{amssymb}
\usepackage{comment}
\usepackage[ruled, linesnumbered]{algorithm2e}
\usepackage{xcolor}
\usepackage[dvipsnames]{xcolor}
\usepackage{booktabs}
\usepackage{pdflscape}
\usepackage{xcolor}
\definecolor{webred}{rgb}{0.5,0,0}
\definecolor{webblue}{rgb}{0,0,0.8}
\usepackage[pdftex,colorlinks,citecolor=webblue,linkcolor=webred,]{hyperref}
\usepackage{orcidlink}  
\usepackage{float}
\usepackage{soul} 
\usepackage{algorithmic} 
\usepackage{algorithm} 

\newcommand{\R}{\mathbb{R}}

\begin{document}
\title{Bridging Neural ODE and ResNet:\\ A Formal Error Bound for Safety Verification}
\author{Abdelrahman Sayed Sayed\orcidlink{0000-0002-8912-0679} \and
Pierre-Jean Meyer\orcidlink{0000-0002-8167-3156} \and
Mohamed Ghazel\orcidlink{0000-0002-1160-7997}}
\authorrunning{A. Sayed et al.}
%
\institute{Univ Gustave Eiffel, COSYS-ESTAS, F-59657 Villeneuve d’Ascq, France
\email{\{abdelrahman.ibrahim,pierre-jean.meyer,mohamed.ghazel\}@univ-eiffel.fr}}

\titlerunning{Bridging Neural ODE and ResNet}

\maketitle              
\begin{abstract}
A neural ordinary differential equation (neural ODE) is a machine learning model that is commonly described as a continuous-depth generalization of a residual network (ResNet) with a single residual block, or conversely, the ResNet can be seen as the Euler discretization of the neural ODE.
These two models are therefore strongly related in a way that the behaviors of either model are considered to be an approximation of the behaviors of the other.
In this work, we establish a more formal relationship between these two models by bounding the approximation error between two such related models.
The obtained error bound then allows us to use one of the models as a verification proxy for the other, without running the verification tools twice: if the reachable output set expanded by the error bound satisfies a safety property on one of the models, this safety property is then guaranteed to be also satisfied on the other model.
This feature is fully reversible, and the initial safety verification can be run indifferently on either of the two models.
This novel approach is illustrated on a numerical example of a fixed-point attractor system modeled as a neural ODE.

\keywords{Neural ODE, ResNet, Formal relationship, Safety verification, Reachability analysis}
\end{abstract}
\section{Introduction}
\label{introduction}

Neural ordinary differential equations (neural ODE) are gaining prominence in continuous-time modeling, offering distinct advantages over traditional neural networks, such as memory efficiency, continuous-time modeling, adaptive computation balancing speed and accuracy~\cite{chen2018neural,kidger2022neuraldifferentialequations,oh2025comprehensive}. This surge in interest stems from recent advancements in differential programming, which have enhanced the ability to model complex dynamics with greater flexibility and precision~\cite{rackauckas2021universaldifferentialequationsscientific}. 

Neural ODE can be viewed as a continuous-depth generalization of residual networks (ResNet)~\cite{he2015deepresiduallearningimage}, and conversely a ResNet represents an Euler discretization of the continuous transformations modeled by a neural ODE~\cite{Haber_2017,lu2020finitelayerneuralnetworks}. 
Unlike ResNet, neural ODE enable smooth and robust representations through continuous dynamics, leading to improved modeling of time-evolving systems~\cite{chen2018neural,Haber_2017}. By interpreting ResNet as discretized neural ODE, we can leverage advanced ODE solvers to enhance computational efficiency and reduce the number of required parameters~\cite{chen2018neural}. Furthermore, the continuous formulation of neural ODE supports flexible handling of varying input resolutions and scales, making them adaptable to diverse data modalities. This perspective also facilitates theoretical analysis using tools from differential equations, providing insights into network stability and convergence~\cite{kidger2022neuraldifferentialequations}.

Despite the growing interest in neural ODE for continuous-time modeling, formal analysis techniques for these models remain underdeveloped~\cite{lopez2022reachabilityanalysisgeneralclass}. Current verification methods for neural ODE are still maturing, with existing reachability approaches primarily focusing on stochastic methods~\cite{gruenbacher2020verificationneuralodesstochastic,gruenbacher2021gotubescalablestochasticverification}. Other works include the NNVODE tool~\cite{lopez2022reachabilityanalysisgeneralclass} which is an extension of the Neural Network Verification (NNV) framework~\cite{tran2020nnvneuralnetworkverification,lopez2023nnv} that investigates reachability for a general class of neural ODE.
Additionally, another line of verification based on topological properties was introduced in~\cite{liang2022safetyverificationneuralnetworks} through a set-boundary method for safety verification of neural ODE and invertible residual networks (i-ResNet)~\cite{behrmann2019invertibleresidualnetworks}. 

\begin{figure}[htb]
    \centering
    \includegraphics[width=\columnwidth]{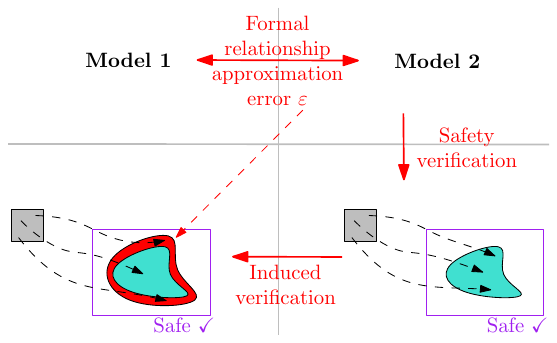}
    \caption{Illustration of the proposed framework to verify Model $1$ based on the outcome of the verification of Model $2$ and a bound $\varepsilon$ on the maximal error between the models.}
    \label{fig:framework}
\end{figure}

The similarity between the neural ODE and ResNet models enables bidirectional safety verification, where the properties verified for one model can be used to deduce safety guarantees for the other one. This motivates our work, which investigates how verification results from one model can serve as a proxy for the other, addressing practical scenarios where only one model or compatible verification tools are available.
The main contributions of this work are as follows:
\renewcommand{\labelitemi}{$\bullet$}
\begin{itemize}
    \item We derive a rigorous bound on the approximation error between the neural ODE and ResNet models for a given input set.
    \item We use the derived error bound in conjunction with the reachable set of one model as a proxy to verify safety properties of the other model, without applying any verification tools to the other model as illustrated in Figure~\ref{fig:framework}.
\end{itemize}

\paragraph{Related work.}  
Although the similarity between the ResNet and neural ODE models is well established~\cite{chen2018neural,kidger2022neuraldifferentialequations}, to the best of our knowledge, very few works have tried connecting these models through some more formal relationships.
These include various theoretical perspectives, such as quantifying the deviation between the hidden state trajectory of a ResNet and its corresponding neural ODE, focusing on approximation error~\cite{sander2022residualneuralnetworksdiscretize}, while~\cite{marion2023generalizationboundsneuralordinary} derives generalization bounds for neural ODE and ResNet using a Lipschitz-based argument, emphasizing the impact of successive weight matrix differences on generalization capability. On the other hand,~\cite{marion2024implicitregularizationdeepresidual} investigates implicit regularization effects in deep ResNet and its impact on training outcomes. While these studies focus on theoretical analyses of approximation error, generalization, and regularization to understand model behavior and performance, our work leverages this relationship for formal safety verification. We propose a verification proxy approach that uses the reachable set of one model to verify the safety properties of the other, incorporating an error bound to ensure conservative over-approximations, which enables practical verification of nonlinear systems.

Abstraction-based verification (i.e.,\ verifying properties of one model by working on an abstraction of its behaviors into a simpler model) has been a popular topic in the past decades outside of the AI field~\cite{tabuada2009verification}. Within the field of AI verification, its primary application has been on abstracting specific model components rather than the whole model itself, as in approaches based on convex relaxation of nonlinear ReLU activation functions~\cite{katz2017reluplex,huang2017safety}.
On the other hand, full-model abstraction has been mostly unexplored for AI verification, except on the topic of neural network model reduction, where the verification of a neural network is achieved at a lower computational cost on a reduced network with less neurons, see e.g.~\cite{boudardara2023innabstract} for unidirectional relationships, or~\cite{xiang2022approximatebisimulationrelationsneural} for bidirectional ones through the use of approximate bisimulation relations.
Although the overall principle of the proposed approach in our paper is similar (abstracting a model by one that over-approximates the set of all its behaviors), the main difference with the above works between two discrete neural networks is that our paper considers the formal relationships between a continuous neural ODE model and a discrete ResNet one.

\paragraph{Organization of the paper.}
The remainder of the paper is structured as follows. First, we formulate the safety verification problem of interest and provide some preliminaries in Section~\ref{section:preliminaries}. In Section~\ref{section:error}, we describe our proposed approach to bound the approximation error between the ResNet and neural ODE models, and use this error bound to verify the safety of one model based on the reachability analysis of the other. Following this, we provide numerical illustrations of our error bounding and verification proxy results (in both directions: from ResNet to neural ODE, and from neural ODE to ResNet) on an academic example in Section~\ref{section:expirements}. Finally, we summarize the main findings of the paper and discuss potential future work in Section~\ref{section:conclusion}.

\section{Preliminaries}
\label{section:preliminaries}

\subsection{Neural ODE and ResNet models}
\label{subsect:models}

We consider the following neural ODE:
\begin{equation}
\dot x(t)=\frac{dx(t)}{dt}=f(x(t)),
\label{eq:nODE}
\end{equation}
with state $x\in\R^n$, initial state $x(0)=u$, and vector field $f:\R^n\rightarrow\R^n$ defined as a finite sequence of classical neural network layers (such as fully connected layers, convolutional layers, activation functions, batch normalization).
The state trajectories of \eqref{eq:nODE} are defined based on the solution $\Phi:\R\times\R^n\rightarrow\R^n$ of the corresponding initial value problem:
$$x(t)=\Phi(t,x(0))=\Phi(t,u).$$

In~\cite{chen2018neural}, such a neural ODE is described as a continuous-depth generalization of a residual neural network constituted of a single residual block.
Conversely, this ResNet can be seen as the Euler discretization of the neural ODE \eqref{eq:nODE}:
\begin{equation}
y=u+f(u),
\label{eq:ResNet}
\end{equation}
where $u\in\R^n$ is the input, $y\in\R^n$ is the output, and the residual function $f:\R^n\rightarrow\R^n$ is identical to the vector field of the neural ODE~\eqref{eq:nODE}.

Since the approach proposed in this paper relies on the Taylor expansion of the trajectories of~\eqref{eq:nODE} up to the second order, we assume here for simplicity that the neural network described by the vector field $f$ is continuously differentiable.

\begin{remark}
The case where $f$ contains piecewise-affine activation functions such as ReLU can theoretically be handled as well, since our approach only really requires their derivatives to be bounded (but not necessarily continuous).
But for the sake of clarity of presentation (to avoid the case decompositions of each ReLU activation), this case is kept out of the scope of the present paper.
\end{remark}

\subsection{Problem definition}
\label{subsect:Prob def}

As mentioned above and in~\cite{chen2018neural}, both the neural ODE and ResNet models describe a very similar behavior, and either model could be seen as an approximation of the other.
Our goal in this paper is to provide a formal comparison of these models in the context of safety verification, by evaluating the approximation error between them.
For such comparison to be meaningful, we consider the outputs $y$ of the ResNet~\eqref{eq:ResNet} on one side, and the outputs $\Phi(1,u)$ of the neural ODE~\eqref{eq:nODE} at continuous depth $t=1$ on the other side, since other values $t\neq 1$ of this continuous depth have no elements of comparison in the discrete architecture of the ResNet. 

Given an initial set $\mathcal{X}_{in}\subseteq\R^n$ for the neural ODE (or equivalently referred to as \emph{input set} for the ResNet), we first define the sets of reachable outputs for either model:
\[\mathcal{R}_{\text{neural ODE}}(\mathcal{X}_{in})=\{y\in \mathbb{R}^n \mid y= \Phi(1,u), \ u\in \mathcal{X}_{in}\},\]

\[\mathcal{R}_{\text{ResNet}}(\mathcal{X}_{in})=\{y\in \mathbb{R}^n \mid y= u + f(u), \ u\in \mathcal{X}_{in}\}.\]

Since we usually cannot compute these output reachable sets exactly, we will often rely on computing an over-approximation denoted as $\Omega(\mathcal{X}_{in})$ such that $\mathcal{R}(\mathcal{X}_{in})\subseteq \Omega(\mathcal{X}_{in})$.

Our first objective is to bound the approximation error between the two models, as formalized below.

\begin{problem}[Error Bounding]
\label{def:error bound}
Given an input set $\mathcal{X}_{in}\subseteq\R^n$, we want to over-approximate the set $\mathcal{R}_\varepsilon(\mathcal{X}_{in})$ of errors between the ResNet~\eqref{eq:ResNet} and neural ODE~\eqref{eq:nODE} models, defined as:
$$\mathcal{R}_\varepsilon(\mathcal{X}_{in})=\left\{\Phi(1, u)-(u+f(u))~|~u\in\mathcal{X}_{in}\right\}.$$
\end{problem}

Our second problem of interest is to use one of our models as a verification proxy for the other.
In other words, we want to combine this error bound with the reachable set of one model to verify the satisfaction of a safety property on the other model, without having to compute the reachable output set of this second model.

\begin{problem}[Verification Proxy]
\label{def:safety verification}
Given an input-output safety property defined by an input set $\mathcal{X}_{in} \subseteq \mathbb{R}^n$ and a safe output set $\mathcal{X}_{s} \subseteq \mathbb{R}^n$, the verification problem consists in checking whether the reachable output set of a model is fully contained in the targeted safe set: $\mathcal{R}(\mathcal{X}_{in})\subseteq\mathcal{X}_{s}$.
In this paper, we want to verify this safety property on one model by relying only on the error set $\mathcal{R}_\varepsilon(\mathcal{X}_{in})$ from Problem~\ref{def:error bound} and the reachability analysis of the other model.
\end{problem}

\section{Proposed approach}
\label{section:error}

As mentioned in Section~\ref{subsect:Prob def}, the ResNet model in~\eqref{eq:ResNet} can be seen as the Euler discretization of the neural ODE~\eqref{eq:nODE} evaluated at continuous depth $t=1$:
\begin{equation}
\label{eq:error}
x(1)=\Phi(1,u)\approx u + f(u)=y.
\end{equation}

Our initial goal, related to Problem~\ref{def:error bound}, is to evaluate this approximation error for a given set of inputs $u\in\mathcal{X}_{in}$.
This is done below through the use of a Taylor expansion and its Lagrange-remainder form, combined later with some tools dedicated for reachability analysis.

\subsection{Lagrange remainder}
\label{subsect:lagrange}

The Taylor expansion of the state trajectory $x(t)$ of the neural ODE~\eqref{eq:nODE} at $t=0$ is given by the infinite sum:
\begin{equation}
x(t)=x(0)+t\frac{dx(0)}{dt}+\frac{t^2}{2!}\frac{d^2x(0)}{dt^2}+\frac{t^3}{3!}\frac{d^3x(0)}{dt^3}+\dots
\label{eq:taylor}
\end{equation}

The Lagrange remainder theorem offers the possibility to truncate~\eqref{eq:taylor} without approximation error, hence preserving the above equality.
We only state below the result in the case of a truncation at the Taylor order $2$ corresponding to the case of interest in our work.

\begin{proposition}[Lagrange remainder~\cite{rudin1976principles}]
\label{prop:lagrange}
There exists $t^*\in[0,t]$ such that
\begin{equation}
x(t)=x(0)+t\frac{dx(0)}{dt}+\frac{t^2}{2!}\frac{d^2x(t^*)}{dt^2}
\label{eq:lagrange}
\end{equation}
\end{proposition}

Notice that in~\eqref{eq:lagrange}, the second order derivative $\frac{d^2x}{dt^2}$ is evaluated at $t^*\in[0,t]$ instead of $t$ as in the Taylor series~\eqref{eq:taylor}.
Although the truncation in Proposition~\ref{prop:lagrange} provides a much more manageable expression than the infinite sum in~\eqref{eq:taylor}, the main difficulty is that this result only states the existence of a $t^*\in[0,t]$ satisfying the equality in~\eqref{eq:lagrange}, but its actual value is unknown.

\subsection{Error function}
\label{subsect:lagrange evaluation}

To compare the continuous state $x(t)$ with the discrete output of the ResNet, the state of the neural ODE~\eqref{eq:nODE} should be evaluated at depth $t=1$.

The first term of the right-hand side in~\eqref{eq:lagrange} is the known initial condition of the neural ODE~\eqref{eq:nODE}: $x(0)=u$.

The second term is provided by the definition of the vector field of the neural ODE~\eqref{eq:nODE}, and thus reduces to:
$$t\frac{dx(0)}{dt}=1\cdot f(x(0))=f(u).$$

The second derivative appearing in the third term of~\eqref{eq:lagrange} can be computed using the chain rule as follows:

\begin{align*}
\frac{d^2x(t)}{dt^2} 
&= \frac{df(x(t))}{dt}\\
&= \frac{\partial f(x(t))}{\partial t}+\frac{\partial f(x(t))}{\partial x}\frac{dx(t)}{dt}\\
&= \frac{\partial f(x(t))}{\partial t}+f'(x(t))f(x(t)).
\end{align*}

In our context of Section~\ref{section:preliminaries}, the function $f$ is assumed not to be explicitly dependent on the depth $t$ due to its definition as a single residual block with classical layers.
Therefore, the partial derivative $\frac{\partial f(x(t))}{\partial t}$ is equal to $0$, and the third term of~\eqref{eq:lagrange} thus reduces to:
$$\frac{t^2}{2!}\frac{d^2x(t^*)}{dt^2}=\frac{1}{2}f'(x(t^*))f(x(t^*)).$$

We can thus re-write~\eqref{eq:lagrange} as an equation defining the output of the neural ODE based on the output of the ResNet (for the same initial state/input $u$) and an error term:
\begin{equation}
    \Phi(1,u)= (u + f(u))+\varepsilon(u),
\label{eq:nODE~ResNet}
\end{equation}
where the approximation error between our models for this particular input $u$ is expressed by the Lagrange remainder of Taylor order 2:
\begin{equation}
    \varepsilon(u)=\frac{1}{2}f'(x(t^*))f(x(t^*)),
\label{eq:error function}
\end{equation}
with $x(t^*)=\Phi(t^*,u)$ for a fixed but unknown $t^*\in[0,1]$.

Equation~\eqref{eq:nODE~ResNet} can also be modified to rather express the outputs of the ResNet based on those of the neural ODE:
\begin{equation}
    u + f(u)=\Phi(1,u)-\varepsilon(u).
\label{eq:ResNet~nODE}
\end{equation}

The error function $\varepsilon:\R^n\rightarrow\R^n$ appearing positively in~\eqref{eq:nODE~ResNet} and negatively in~\eqref{eq:ResNet~nODE} is defined in~\eqref{eq:error function} only for a specific input $u$.
However, in the context of our Problem~\ref{def:error bound}, we are interested in analyzing the approximation error between both models over an input set $\mathcal{X}_{in} \subseteq \mathbb{R}^n$.
In addition, since the specific value of $t^*$ is unknown, we need to bound~\eqref{eq:error function} for any possible value of $t^* \in [0,1]$.
Therefore in the next sections, we focus on converting the equalities~\eqref{eq:nODE~ResNet}-\eqref{eq:ResNet~nODE} to set inclusions over all $u\in\mathcal{X}_{in}$ and $t^* \in [0,1]$.

\subsection{Bounding the error set}
\label{subsect:bounding error}

The reachable error set $\mathcal{R}_\varepsilon(\mathcal{X}_{in})$ introduced in Problem~\ref{def:error bound}, can be redefined based on the error function~\eqref{eq:error function} as follows:
\begin{align}
\mathcal{R}_\varepsilon(\mathcal{X}_{in})&=\left\{\Phi(1, u)-(u+f(u))~|~u\in\mathcal{X}_{in}\right\}\nonumber\\
&=\left\{\left.\frac{1}{2}f'(\Phi(t^*,u))f(\Phi(t^*,u))~\right|~t^* \in [0,1],~u\in\mathcal{X}_{in}\right\}.
\label{eq:error set 1}
\end{align}
To solve Problem~\ref{def:error bound}, our objective is thus to compute an over-approximation $\Omega_\varepsilon(\mathcal{X}_{in})$ bounding the error set: $\mathcal{R}_\varepsilon(\mathcal{X}_{in})\subseteq\Omega_\varepsilon(\mathcal{X}_{in})$.

The first step (corresponding to line 1 in Algorithm~\ref{alg:nODE}) is to compute the reachable tube of all possible states that can be reached by the neural ODE~\eqref{eq:nODE} over the whole range $t\in[0,1]$ and for any initial state $x(0)=u\in\mathcal{X}_{in}$.
This reachable tube can be defined similarly to $\mathcal{R}_{\text{neural ODE}}(\mathcal{X}_{in})$ in Section~\ref{subsect:Prob def} but for all possible depth $t\in[0,1]$ instead of only the final one:
\begin{equation*}
\label{eq:reach tube}
\mathcal{R}^{\text{tube}}_{\text{neural ODE}}(\mathcal{X}_{in})=\{\Phi(t,u)\in \mathbb{R}^n \mid t\in[0,1],~u\in \mathcal{X}_{in}\}.
\end{equation*}
Since in most cases this set cannot be computed exactly, we instead use off-the-shelf reachability analysis toolboxes to compute an over-approximating set $\Omega^{\text{tube}}_{\text{neural ODE}}(\mathcal{X}_{in})$ such that $\mathcal{R}^{\text{tube}}_{\text{neural ODE}}(\mathcal{X}_{in})\subseteq\Omega^{\text{tube}}_{\text{neural ODE}}(\mathcal{X}_{in})$.

The error set can then be re-written based on the above reachable tube definition, by replacing $\Phi(t^*,u)$ (with $t^* \in [0,1]$ and $u \in \mathcal{X}_{in}$) in~\eqref{eq:error set 1} by $x\in\mathcal{R}^{\text{tube}}_{\text{neural ODE}}(\mathcal{X}_{in})$.

\begin{align}
\mathcal{R}_\varepsilon(\mathcal{X}_{in})
&=\left\{\left.\frac{1}{2}f'(x)f(x)~\right|~x\in\mathcal{R}^{\text{tube}}_{\text{neural ODE}}(\mathcal{X}_{in})\right\}\nonumber\\
&\subseteq\left\{\left.\frac{1}{2}f'(x)f(x)~\right|~x\in\Omega^{\text{tube}}_{\text{neural ODE}}(\mathcal{X}_{in})\right\}.
\label{eq:error set 2}
\end{align}

The next step, in line 2 of Algorithm~\ref{alg:nODE}, is to over-approximate this error set $\mathcal{R}_\varepsilon(\mathcal{X}_{in})$.
One possible approach to achieve this is to define the static function $\varepsilon=\frac{1}{2}f'(x)f(x)$ and apply to it some set-propagation techniques (such as interval arithmetic~\cite{jaulin2001interval}, Taylor models~\cite{makino2003taylor}, or affine arithmetic~\cite{de2004affine}) to bound the set of output errors $\varepsilon$ corresponding to any state $x\in\Omega^{\text{tube}}_{\text{neural ODE}}(\mathcal{X}_{in})$ in the reachable tube over-approximation.
An alternative approach, which provided a tighter error bounding set in the particular case of the numerical example presented in Section~\ref{section:expirements}, is to define the discrete-time nonlinear system $x^+=\frac{1}{2}f'(x)f(x)$, and then use existing reachability analysis toolboxes to over-approximate the reachable set of this system after one time step, which corresponds to bounding the image of the error function.
Note that in this case, it is important that this final reachable set is computed as a single step, and not decomposed into a sequence of smaller intermediate steps whose iterative updates of the internal state would have no mathematical meaning for the static (stateless) error function.

As a consequence of the equalities and set inclusions in~\eqref{eq:error set 1}-\eqref{eq:error set 2} and the fact that the reachability methods to be used in the first two steps of Algorithm~\ref{alg:nODE} described above guarantee that the obtained sets are over-approximations of the output or reachable sets of interest, we have thus reached a solution to Problem~\ref{def:error bound}.
\begin{theorem}
\label{thm:error}
The set $\Omega_\varepsilon(\mathcal{X}_{in})$ obtained after applying this second step described above solves Problem~\ref{def:error bound}:
$$\mathcal{R}_\varepsilon(\mathcal{X}_{in})=\left\{\Phi(1, u)-(u+f(u))~|~u\in\mathcal{X}_{in}\right\}\subseteq\Omega_\varepsilon(\mathcal{X}_{in}).$$
\end{theorem}

Note that the error bound in Theorem~\ref{thm:error} is defined as a set in the state space of the neural ODE.
This differs from the approach in~\cite{sander2022residualneuralnetworksdiscretize}, where the error bound is defined as a positive scalar.

A second and more important difference with this work is the tightness of the obtained error bounds.
Indeed, if we adapt the results from~\cite{sander2022residualneuralnetworksdiscretize} to the context of our framework described in Section~\ref{section:preliminaries}, their error bound is expressed as: 
$$\varepsilon\leq \frac{e^L-1}{L}\left\|\frac{1}{2}f'(x)f(x)\right\|_\infty,~\forall x\in\mathcal{R}^{\text{tube}}_{\text{neural ODE}}(\mathcal{X}_{in}),$$
where $L$ is a Lipschitz constant of the neural ODE vector field.
The term $\left\|\frac{1}{2}f'(x)f(x)\right\|_\infty$ can be obtained by first over-approximating the error set by $\Omega_\varepsilon(\mathcal{X}_{in})$ in the same way we did, but the infinity norm forces to expand this set to make it symmetrical around $0$, and then keeping only the maximum value among its components (thus corresponding to a second expansion of this set into an hypercube whose width along all dimensions is the largest width of the previous set).
In addition, for any system with non-zero Lipschitz constant, the factor $\frac{e^L-1}{L}$ is always greater than $1$, which increases this error bound even more.

In summary, this scalar error bound is doubly more conservative than our proposed set-based error bound.
The comparison of both approaches is illustrated in the numerical example of Section~\ref{section:expirements}.

\subsection{Verification proxy}

To address Problem~\ref{def:safety verification}, we leverage the similar behavior between the neural ODE and ResNet models to verify safety properties on one model using the reachable set of the other, combined with the error bound from Theorem~\ref{thm:error}. Specifically, we want to verify whether the reachable output set of a model is contained in the safe set $\mathcal{X}_{s}$, i.e., $\mathcal{R}(\mathcal{X}_{in}) \subseteq \mathcal{X}_s$. 

We first focus on the case of Algorithm~\ref{alg:nODE} to verify the safety property on the neural ODE, based on the reachability analysis of the ResNet.
This first verification proxy relies on the set-based version of~\eqref{eq:nODE~ResNet} using the Minkowski sum:
\begin{equation}
    \mathcal{R}_{\text{neural ODE}}(\mathcal{X}_{in}) \subseteq \Omega_{\text{ResNet}}(\mathcal{X}_{in}) + \Omega_{\varepsilon}(\mathcal{X}_{in}),
    \label{eq:nODE~ResNet-sb}
\end{equation}
stating that the reachable output set of the neural ODE is contained in the output set over-approximation of the ResNet $\Omega_{\text{ResNet}}(\mathcal{X}_{in})$, expanded by the bounding set of the error $\Omega_{\varepsilon}(\mathcal{X}_{in})$ obtained after applying the first two lines of Algorithm~\ref{alg:nODE} as described in Section~\ref{subsect:bounding error}.

Therefore, this verification procedure is achieved as in Algorithm~\ref{alg:nODE}, by first using existing set-propagation or reachability analysis tools to compute an over-approximation $\textcolor{blue}{\Omega_{\text{ResNet}}(\mathcal{X}_{in})}$ of the ResNet output set (line 3).
Then in line 4, an over-approximation of the neural ODE output set can be deduced from~\eqref{eq:nODE~ResNet-sb} by taking the Minkowski sum of $\textcolor{blue}{\Omega_{\text{ResNet}}(\mathcal{X}_{in})}$ and our error bound $\textcolor{red}{\Omega_{\varepsilon}(\mathcal{X}_{in})}$.
If $\Omega_{\text{neural ODE}}(\mathcal{X}_{in})$ is contained in the safe set $\textcolor{OliveGreen}{\mathcal{X}_{s}}$, then the neural ODE satisfies the safety property, otherwise the result is inconclusive (line 5-9).

\begin{algorithm}[htb]
\caption{Safety Verification Framework for neural ODE based on ResNet}
\label{alg:nODE}
\textbf{Input}: a neural ODE, an input set $\mathcal{X}_{in}$ and a safe set $\mathcal{X}_s$.\\
\textbf{Output}: \textbf{Safe} or \textbf{Unknown}.
\begin{algorithmic}[1] 
\STATE compute an over-approximation of the reachable tube of the neural ODE $\Omega^{\text{tube}}_{\text{neural ODE}}(\mathcal{X}_{in})$;
\STATE compute the over-approximation of the error set $\textcolor{red}{\Omega_{\varepsilon}(\mathcal{X}_{in})}$, $\forall x \in \Omega^{\text{tube}}_{\text{neural ODE}}(\mathcal{X}_{in})$;
\STATE compute the over-approximation of the ResNet output $\textcolor{blue}{\Omega_{\text{ResNet}}(\mathcal{X}_{in})}$;
\STATE deduce an over-approximation of the neural ODE output\\ $\Omega_{\text{neural ODE}}(\mathcal{X}_{in}) = \textcolor{red}{\Omega_{\text{ResNet}}(\mathcal{X}_{in})+\Omega_{\varepsilon}(\mathcal{X}_{in})}$;
\IF {$\Omega_{\text{neural ODE}}(\mathcal{X}_{in}) \subseteq \textcolor{OliveGreen}{\mathcal{X}_{s}}$}
\STATE return \textbf{Safe}
\ELSE
\STATE return \textbf{Unknown}
\ENDIF
\end{algorithmic}
\end{algorithm}

\bigskip
Reversing the roles, the case of verifying the ResNet based on the reachability analysis of the neural ODE is described in Algorithm~\ref{alg:ResNet}.
This case is very similar to the previous one, so we focus here on the main differences with Algorithm~\ref{alg:nODE}.
The first difference is that in~\eqref{eq:ResNet~nODE}, the term representing the approximation error between the models appears with a negative sign.
Therefore, when converting this equation into a set inclusion similarly to~\eqref{eq:nODE~ResNet-sb}, we need to be careful to add the negation of the error set (and not to do a set difference, which is not the correct set operation in our case).
We thus introduce the negative error set 
$$\Omega_{-\varepsilon}(\mathcal{X}_{in}) = \{-\varepsilon \mid \varepsilon \in \Omega_\varepsilon(\mathcal{X}_{in})\},$$
in order to convert~\eqref{eq:ResNet~nODE} into its set-based notation as follows:
\begin{equation}
    \mathcal{R}_{\text{ResNet}}(\mathcal{X}_{in}) \subseteq \Omega_{\text{neural ODE}}(\mathcal{X}_{in}) + \Omega_{-\varepsilon}(\mathcal{X}_{in}).
    \label{eq:ResNet~nODE-sb}
\end{equation}

The second difference is that in line 3 of Algorithm~\ref{alg:ResNet}, we compute an over-approximation of the reachable set of the neural ODE, using any classical tools for reachability analysis of continuous-time nonlinear systems, and add it to the negative error set to obtain an over-approximation of the ResNet output set.
This final set can then similarly be used to verify the satisfaction of the safety property on the ResNet model.

\begin{algorithm}[htb]
\caption{Safety Verification Framework for ResNet based on neural ODE}
\label{alg:ResNet}
\textbf{Input}: a ResNet, an input set $\mathcal{X}_{in}$ and a safe set $\mathcal{X}_s$.\\
\textbf{Output}: \textbf{Safe} or \textbf{Unknown}.
\begin{algorithmic}[1] 
\STATE compute an over-approximation of the reachable tube of the neural ODE $\Omega^{\text{tube}}_{\text{neural ODE}}(\mathcal{X}_{in})$;
\STATE compute the over-approximation of the negative error set $\textcolor{red}{\Omega_{-\varepsilon}(\mathcal{X}_{in})}$, $\forall x \in \Omega^{\text{tube}}_{\text{neural ODE}}(\mathcal{X}_{in})$;
\STATE compute the over-approximation of the neural ODE output $\Omega_{\text{neural ODE}}(\mathcal{X}_{in})$;
\STATE deduce an over-approximation of the ResNet output\\ $\textcolor{blue}{\Omega_{\text{ResNet}}(\mathcal{X}_{in})} = \textcolor{red}{\Omega_{\text{neural ODE}}(\mathcal{X}_{in})+\Omega_{-\varepsilon}(\mathcal{X}_{in})}$;
\IF {$\textcolor{blue}{\Omega_{\text{ResNet}}(\mathcal{X}_{in})} \subseteq \textcolor{OliveGreen}{\mathcal{X}_{s}}$}
\STATE return \textbf{Safe}
\ELSE
\STATE return \textbf{Unknown}
\ENDIF
\end{algorithmic}
\end{algorithm}

\begin{theorem}[Soundness]
\label{sound}
For the case that either Algorithm~\ref{alg:nODE} or~\ref{alg:ResNet} returns \textbf{\em Safe},  the safety property in the sense of Problem~\ref{def:safety verification} holds true\em~\cite{liang2022safetyverificationneuralnetworks}.  
\end{theorem}

The soundness of the verification framework is guaranteed because both algorithms rely on over-approximations of the true reachable sets. Specifically,~\eqref{eq:nODE~ResNet-sb} ensures that $\mathcal{R}_{\text{neural ODE}}(\mathcal{X}_{in}) \subseteq \Omega_{\text{neural ODE}}(\mathcal{X}_{in})$, and~\eqref{eq:ResNet~nODE-sb} ensures $\textcolor{blue}{\mathcal{R}_{\text{ResNet}}(\mathcal{X}_{in})} \subseteq \textcolor{blue}{\Omega_{\text{ResNet}}(\mathcal{X}_{in})}$. These inclusions hold due
to the conservative nature of the considered reachability analysis and error bound computations in Section~\ref{subsect:bounding error} (Theorem~\ref{thm:error}).

\section{Numerical illustration}
\label{section:expirements}

In this section, a commonly used neural ODE academic example~\cite{lopez2023nnv,lopez2022reachabilityanalysisgeneralclass} is used to demonstrate the verification proxy between the two models, which is the Fixed-Point Attractor (FPA)~\cite{musau2018continuous} that consists of one nonlinear neural ODE. 

\noindent \textbf{Experiment Setting:} All the experiments\footnote{Code available in the following repository: \url{https://github.com/ab-sayed/Formal-Error-Bound-for-Safety-Verification-of-neural-ODE}} herein are run on MATLAB 2024b with Continuous Reachability Analyzer (CORA) version 2024.4.0 with an Intel (R) Core (TM) i5-1145G7 CPU@2.60 GHz and 32 GB of RAM. 

\subsection{System description}
\label{subsect:system desc}

The FPA system is a nonlinear dynamical system with dynamics that converge to a fixed point (an equilibrium state) under certain conditions~\cite{beer1995dynamics}, and the fixed-point aspect makes it a useful model for studying convergence and stability, which are important in safety-critical applications where the system must not diverge or enter unsafe states.
As in the proposed benchmark in~\cite{musau2018continuous}, we consider here the following $5$-dimensional neural ODE approximating the FPA dynamics:
$$\dot{x} = f(x) = \tau x + W \text{tanh}(x),$$
where $x \in \mathbb{R}^5$ is the state vector, $\tau=-10^{-6}$ is a time constant for the neurons, $W \in \mathbb{R}^{5\times5}$ is a composite weight matrix defined as
$W=\begin{pmatrix}0_{2\times 2} & A\\0_{3\times 2} & BA\end{pmatrix}$ with
$A=\begin{pmatrix}-1.20327 & -0.07202 & -0.93635\\1.18810 & -1.50015 & 0.93519\end{pmatrix}$ and
$B=\begin{pmatrix}1.21464 & -0.10502\\ 0.12023 & 0.19387\\ -1.36695 & 0.12201\end{pmatrix}$, and $\text{tanh}(x)$ is the hyperbolic tangent activation function applied element-wise to the state vector $x$. 

We choose our safety property defined by the input set $\mathcal{X}_{in}\approx[0.45,0.55]\times[0.72,0.88]\times[0.47,0.58]\times[0.19,0.24]\times[-0.64,-0.53]$ (its exact numerical values are provided in the code linked below) and the safe set $\mathcal{X}_{s} = [0.2, 0.6] \times [0.3, 0.85] \subset \mathbb{R}^{2}$, that only focuses on the projection of the state onto its first two dimensions, i.e., using an output function $h(x)=(x_{1}, x_{2})$.
In the case of the neural ODE, we thus want to verify that for all initial state $x(0)\in\mathcal{X}_{in}$, we have $h(x(1))\in\mathcal{X}_{s}$.

\subsection{Computing the error bound}
\label{subsect:error bound comp}

\begin{figure}[htb]
    \centering
    \includegraphics[width=\columnwidth]{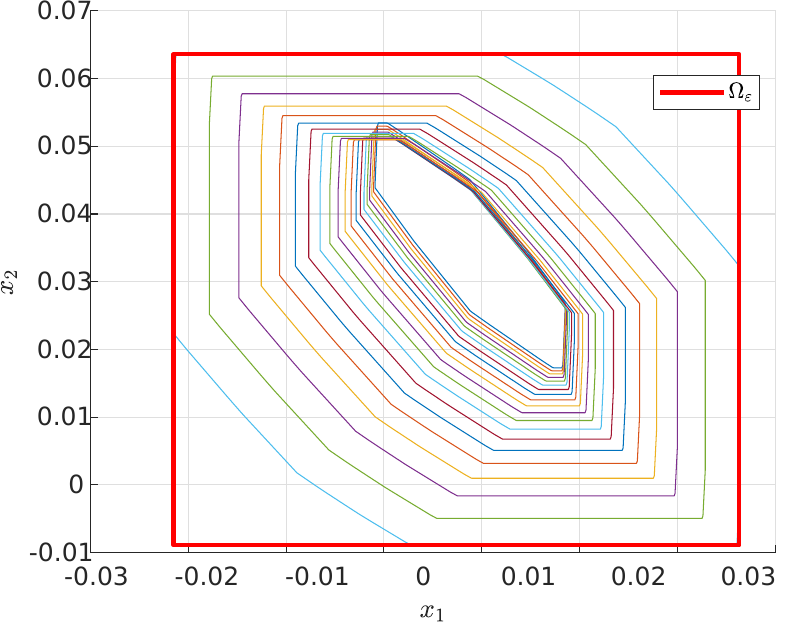}
    \caption{Illustration of the error over-approximation}
    \label{fig:error}
\end{figure}

Using CORA~\cite{althoff2015introduction}, we compute the error bound $\Omega_{\varepsilon}(\mathcal{X}_{in})$ from Theorem~\ref{thm:error} as follows. First, we over-approximate the reachable tube of the neural ODE $\mathcal{R}^{\text{tube}}_{\text{neural ODE}}$ over the time interval $[0,1]$ as a sequence of zonotopes, where each zonotope corresponds to an intermediate time range. For each zonotope in the reachable tube, we bound the image of the error function~\eqref{eq:error function} by applying a discrete-time reachability analysis method at $t=1$. This results in a new zonotope that over-approximates the error set starting from that particular reachable tube zonotope. The total error set is thus guaranteed to be contained in the union of these error zonotopes across all time steps. To simplify its use in the safety verification experiments in Section~\ref{subsect:safety verif}, we compute the interval hull of this union, yielding a hyperrectangle that over-approximates $\textcolor{red}{\Omega_{\varepsilon}(\mathcal{X}_{in})}$ illustrated in Figure~\ref{fig:error} in red, and showing 20 error zonotopes in different colors, corresponding to the error bound of each intermediate time range used in the reachable tube.

\begin{figure}[htb]
    \centering
   \includegraphics[width=\columnwidth]{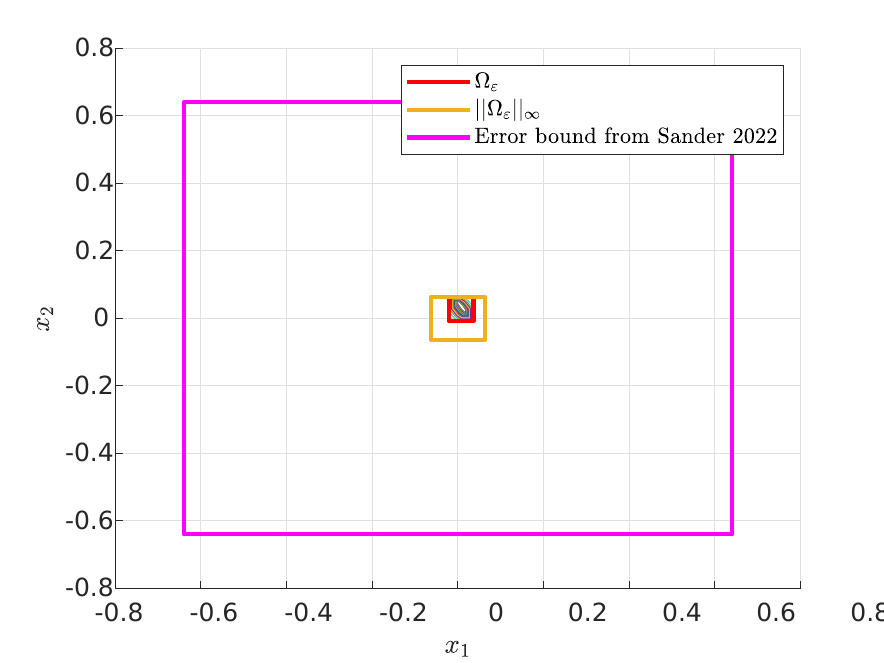}
    \caption{Comparison of the error bounds obtained from our approach in red and the one from~\cite{sander2022residualneuralnetworksdiscretize} in \textcolor{Magenta}{magenta}}
    \label{fig:error bound comparison}
\end{figure}

\bigskip
To contextualize our proposed error bound, we compare it with the error bound proposed in~\cite{sander2022residualneuralnetworksdiscretize}. For that, we first compute the infinity norm of our error set $\|\textcolor{red}{\Omega_{\varepsilon}(\mathcal{X}_{in})}\|_{\infty}=0.064$, which corresponds to a positive and scalar bound on the error, thus implying that its set representation in the state space (represented in yellow in Figure~\ref{fig:error bound comparison}) is necessarily symmetrical around $0$ and with the width that is identical on all dimensions (since the infinity norm takes the largest width across all dimensions).
The set-based error bound ($\textcolor{red}{\Omega_{\varepsilon}(\mathcal{X}_{in})}$ represented in red) obtained from our method is thus always contained in this infinity norm.

Next, we compute the Lipschitz constant for the vector field of the neural ODE $L=\|\tau+W\|_\infty=3.62$, and then we obtain the error bound in~\cite{sander2022residualneuralnetworksdiscretize} as $\frac{(e^{L}-1)}{L} \|\Omega_{\varepsilon}(\mathcal{X}_{in})\|_{\infty}=0.64$. 
This final error bound, represented in \textcolor{Magenta}{magenta} in Figure~\ref{fig:error bound comparison}, is 10 times wider (on each dimension) than the infinity norm of our error set in yellow, and about $16$ millions times larger (in volume over the $5$-dimensional state space) than our error set $\textcolor{red}{\Omega_{\varepsilon}(\mathcal{X}_{in})}$ in red.
The improved tightness of our proposed approach is therefore very significant.

\subsection{Experiments on safety verification}
\label{subsect:safety verif}

Using the error bound computed in Section~\ref{subsect:error bound comp}, we can verify safety properties for the neural ODE output set based on the ResNet output set and the error bound set (i.e., $\textcolor{red}{\Omega_{\text{ResNet}}(\mathcal{X}_{in}) + \Omega_{\varepsilon}(\mathcal{X}_{in})}$), or vice versa for the ResNet output set based on the neural ODE output set and the negative error bound set (i.e., $\textcolor{red}{\Omega_{\text{neural ODE}}(\mathcal{X}_{in}) + \Omega_{-\varepsilon}(\mathcal{X}_{in})}$).

\begin{figure}[htb]
    \centering
    \includegraphics[width=0.9\columnwidth]{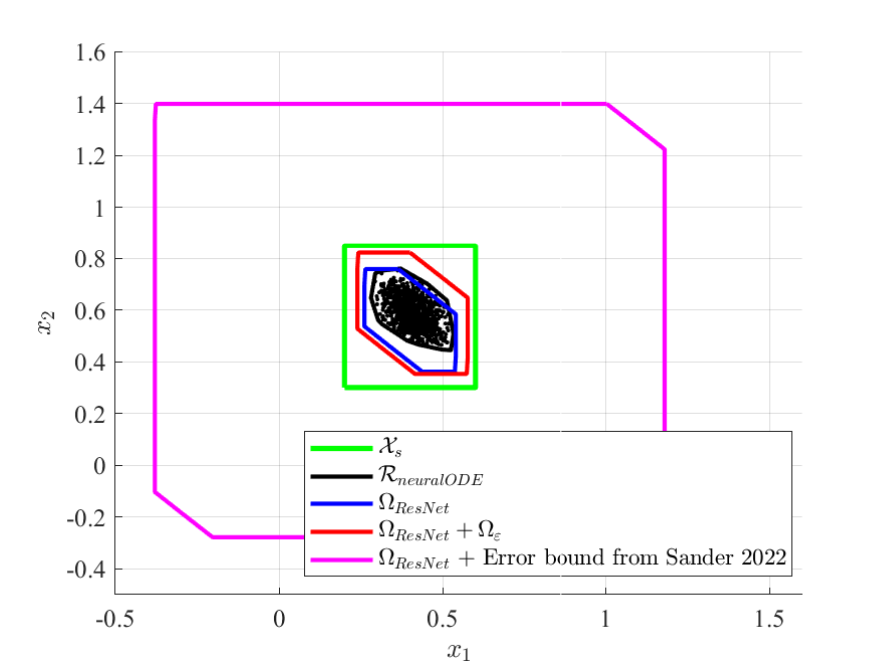}
    \caption{Verification of neural ODE based on ResNet}
    \label{fig:ResNet to nODE}
\end{figure}

In Figure~\ref{fig:ResNet to nODE}, we compute the over-approximation of the ResNet output set $\textcolor{blue}{\Omega_{\text{ResNet}}}$ using simple bound propagation through the ResNet function with CORA. By adding the error bound  $\textcolor{red}{\Omega_{\varepsilon}}$, we obtain a zonotope (shown in \textcolor{red}{red}) that is guaranteed to contain $\mathcal{R}_{\text{neural ODE}}(\mathcal{X}_{in})$. The figure also includes black points representing neural ODE outputs for random initial conditions in $\mathcal{X}_{in}$, with their convex hull (black set) approximating the true reachable set $\mathcal{R}_{\text{neural ODE}}(\mathcal{X}_{in})$. Since the safe set $\textcolor{OliveGreen}{\mathcal{X}_s}$ contains the over-approximation $\textcolor{red}{\Omega_{\text{ResNet}}(\mathcal{X}_{in}) + \Omega_{\varepsilon}(\mathcal{X}_{in})}$, we guarantee that the neural ODE true reachable set is safe, as:
$$\textcolor{OliveGreen}{\mathcal{X}_s}\supseteq\textcolor{red}{\Omega_{\text{ResNet}}(\mathcal{X}_{in}) + \Omega_{\varepsilon}(\mathcal{X}_{in})} \supseteq \mathcal{R}_{\text{neural ODE}}.$$

From Figure~\ref{fig:ResNet to nODE}, we can see that the ResNet and neural ODE reachable sets are very similar due to the ResNet role as a discretization of the neural ODE, but they are not identical. Indeed, some neural ODE outputs (black points) lie outside $\textcolor{blue}{\Omega_{\text{ResNet}}}$, highlighting the necessity of the error bound $\textcolor{red}{\Omega_{\varepsilon}(\mathcal{X}_{in})}$ to ensure that the over-approximation captures all possible neural ODE outputs.

\begin{figure}[htb]
    \centering
    \includegraphics[width=0.9\columnwidth]{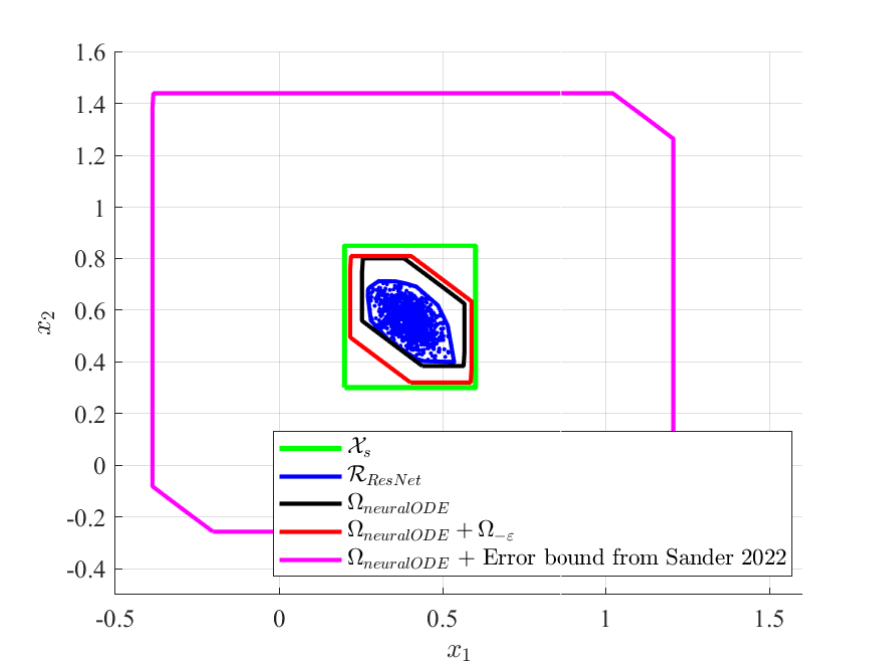}
    \caption{Verification of ResNet based on neural ODE}
    \label{fig:nODE to ResNet}
\end{figure}

Conversely, in Figure~\ref{fig:nODE to ResNet}, we compute the over-approximation of the neural ODE reachable set $\textcolor{black}{\Omega_{\text{neural ODE}}(\mathcal{X}_{in})}$. By adding the negative error bound $\textcolor{red}{\Omega_{-\varepsilon}}$, we obtain a zonotope (shown in \textcolor{red}{red}) that encapsulates $\textcolor{blue}{\mathcal{R}_{\text{ResNet}}(\mathcal{X}_{in})}$. Similarly, the figure includes \textcolor{blue}{blue} points representing ResNet outputs for random inputs in $\mathcal{X}_{in}$, with their convex hull (\textcolor{blue}{blue} set) approximating the true reachable set $\textcolor{blue}{\mathcal{R}_{\text{ResNet}}(\mathcal{X}_{in})}$. Since the safe set $\textcolor{OliveGreen}{\mathcal{X}_s}$ is a super set that contains the over-approximation $\textcolor{red}{\Omega_{\text{neural ODE}}(\mathcal{X}_{in}) + \Omega_{-\varepsilon}(\mathcal{X}_{in})}$, we guarantee that the ResNet true reachable set is safe, as:
$$\textcolor{OliveGreen}{\mathcal{X}_s}\supseteq\textcolor{red}{\Omega_{\text{neural ODE}}(\mathcal{X}_{in}) + \Omega_{-\varepsilon}(\mathcal{X}_{in})} \supseteq \textcolor{blue}{\mathcal{R}_{\text{ResNet}}}.$$

We can also remark that the \textcolor{Magenta}{magenta sets} obtained by adding the error bound proposed in~\cite{sander2022residualneuralnetworksdiscretize} to the ResNet and neural ODE reachable sets in Figures~\ref{fig:ResNet to nODE} and~\ref{fig:nODE to ResNet}, extends significantly beyond the green safe set, preventing us from successfully guaranteeing the safety of the models.

\section{Conclusion}
\label{section:conclusion}

In this paper, we propose a set-based method to bound the error between a neural ODE model and its ResNet approximation.
This approach is based on reachability analysis tools applied to the Lagrange remainder in the Taylor expansion of the neural ODE trajectories, and is shown both theoretically and numerically to provide significantly tighter over-approximation of this approximation error than previous results in~\cite{sander2022residualneuralnetworksdiscretize}.
As the second contribution of this paper, the obtained bounding set of the approximation error between the two models is used to verify a safety property on either of the two models by applying reachability or verification tools only on the other model.
This approach is fully reversible and either model can be used as the verification proxy for the other.
These contributions and their improvement with respect to~\cite{sander2022residualneuralnetworksdiscretize} have been illustrated on a numerical example of a fixed-point attractor system modeled as a neural ODE.

In future works, we plan to explore additional sources of complexity for these approaches, such as handling non-smooth activation functions (e.g.\ ReLU), and the case where the neural ODE vector field is explicitly dependent on the depth variable $t$, thus corresponding to ResNet with multiple residual blocks. Additionally, we aim to study the versatility of this verification proxy approach by applying it to other complex nonlinear dynamical systems or neural network architectures.

\section*{Acknowledgement}
This project has received funding from the European Union’s Horizon 2020 research and innovation programme under the Marie Skłodowska-Curie COFUND grant agreement no. 101034248.

\bibliographystyle{splncs04}
\bibliography{bib}
\end{document}